%% file: main.tex
\documentclass[sigconf, nonacm]{acmart}

\AtBeginDocument{%
  }

\setcopyright{acmlicensed}
\copyrightyear{2018}
\acmYear{2018}
\acmDOI{XXXXXXX.XXXXXXX}
\acmConference[Conference acronym 'XX]{Make sure to enter the correct
  conference title from your rights confirmation email}{June 03--05,
  2018}{Woodstock, NY}
\acmISBN{978-1-4503-XXXX-X/2018/06}

\usepackage{xcolor}
\usepackage{microtype}
\usepackage{algorithm}
\usepackage{algpseudocode}
\usepackage{graphicx}
\usepackage{subcaption}
\usepackage{booktabs}
\usepackage{amsmath}

\usepackage{amssymb}
\usepackage{mathtools}
\usepackage{amsthm}
\usepackage{multicol}
\usepackage{multirow}
\usepackage{hyperref}
\usepackage{pifont}       % \ding{xx}
\usepackage{bbding}       % \Checkmark,\XSolid,... (需要和pifont宏包共同使用)
\usepackage{fontawesome}  % \faCheck,\faTimes
\usepackage{colortbl}     % 表格背景颜色
\usepackage{setspace}
\usepackage[inline]{enumitem}
\begin{document}

\title{\textit{HeiSD}: Hybrid Speculative Decoding for Embodied Vision-Language-Action Models with Kinematic Awareness}

\author{Zihao Zheng$^{1}$, Zhihao Mao$^{2}$, Sicheng Tian$^{3}$, Jiayu Chen$^{1}$, Maoliang Li$^{1}$, Xinhao Sun$^{4}$, Zhaobo Zhang$^{1}$, Xuanzhe Liu$^{1}$, Donggang Cao$^{1}$, Hong Mei$^{1}$, Xiang Chen$^{1}$}
\affiliation{
$^{1}$ School of Computer Science, Peking University \\
$^{2}$ School of Computer Science, China University of Geosciences (Wuhan) \\
$^{3}$ School of Artificial Intelligence, Beijing Normal University \\
$^{4}$ School of EECS, Peking University 
\country{}
}

\input{_tex/0_abstract.tex}

\maketitle
\input{_tex/1_introduction.tex}
\input{_tex/2_background.tex}
\input{_tex/3_analysis.tex}
\input{_tex/4_rsd-optimization.tex}
\input{_tex/5_hybrid.tex}
\input{_tex/6_implementation.tex}
\input{_tex/7_experiments.tex}
\input{_tex/8_conclusion.tex}


% \bibliographystyle{ACM-Reference-Format}
% \bibliography{ref/reference.bib}

% \clearpage
\appendix
\onecolumn
\input{_tex/9_appendix.tex}

\twocolumn
\clearpage
\bibliographystyle{ACM-Reference-Format}
\bibliography{ref/reference.bib}

% \newpage
% \appendix
% \onecolumn
% \input{_tex/9_appendix.tex}

\end{document}

%% file: _tex/0_abstract.tex
\begin{abstract}
Vision-Language-Action (VLA) Models have become the mainstream solution for robot control, but suffer from slow inference speeds.
Speculative Decoding (SD) is a promising acceleration method which can be divided into two categories: drafter-based SD and retrieval-based SD.
Each of the two methods demonstrates complementary advantages and limitations when applied to VLA models, leading to the hypothesis that a hybrid approach integrating these two methods will yield better performance.
In this paper, we first conduct a series of detailed analyses to reveal the advantages and feasibility of hybrid utilization.
However, even with the aforementioned key insights, implementing hybrid SD in VLA models presents several challenges:
(1) draft rejection and persistent errors in retrieval-based SD; (2) difficulty in determining the hybrid boundary.
To address these, we propose the \textit{HeiSD} framework.
We propose a retrieval-based SD optimization method in \textit{HeiSD}, which contains a verify-skip mechanism and a sequence-wise relaxed acceptance strategy.
Moreover, we proposed a kinematic-based fused metric in \textit{HeiSD} to automatically determine the hybrid boundary.
Experimental results demonstrate that \textit{HeiSD} attains a speedup of up to 2.45$\times$ in simulation benchmarks and 2.06$\times$$\sim$2.41$\times$ in real-world scenarios, while sustaining a high task success rate.

\end{abstract}

%% file: _tex/1_introduction.tex
\section{Introduction}
\label{tex:introduction}

% =======================================================================
% Paragraph 1: VLA是什么
% =======================================================================
Vision-Language-Action (VLA) models have emerged as the mainstream solution for Embodied Intelligence~\cite{survey-1, survey-2}.
A VLA model typically consists of three components: visual and text encoders, a Large Language Model (LLM), and an action decoder.
Leveraging the comprehension and reasoning capabilities of LLMs, VLA models enables accurate action generation~\cite{openvla, rt2}.

% =======================================================================
% Paragraph 2: VLA的计算困境+加速SOTA（SD是runtime中的一环）
% =======================================================================
Despite their impressive performance, the heavy computational demands of VLA models limit their inference speed, preventing them from meeting real-time requirements~\cite{realtimevla, efficientvla}.
To boost the inference speed of VLA models, existing work has integrated large-model inference optimization techniques into VLAs, covering: model architecture innovation~\cite{robomamba, tinyvla, edgevla, fast}, model compression~\cite{qvla, dyq-vla, QAIL, moqa}, runtime optimization~\cite{molevla, ceedvla, deervla} and deployment design~\cite{RAPID, roboecc}.
In such runtime optimization methods, Speculative Decoding (SD)~\cite{specvla, kerv} is a promising method that can accelerate VLA models' inference to meet real-time requirements.

% =======================================================================
% Paragraph 2: SD有两种（drafter-based SD and retrieval-based SD）
% =======================================================================
The essence of SD lies in employing low-cost methods to rapidly generate token sequences. 
These sequences then undergo parallel verification and selective acceptance by Large Language Models (LLMs), thereby enhancing inference speed.
Existing SD methods can be categorized into two types: drafter-based SD~\cite{speculative, decspecdec} and retrieval-based SD~\cite{HierarchicalDrafting, REST, PROMTEC}.
Drafter-based SD leverages a small model (trained from scratch or fine-tuned) to generate draft token sequences.
In contrast, retrieval-based SD does not rely on a dedicated draft model; instead, it uses a prebuilt vector database and retrieves relevant content from it to obtain draft token sequences.

\begin{figure*}[!t]
  \begin{center}
    \centerline{\includegraphics[width=7in]{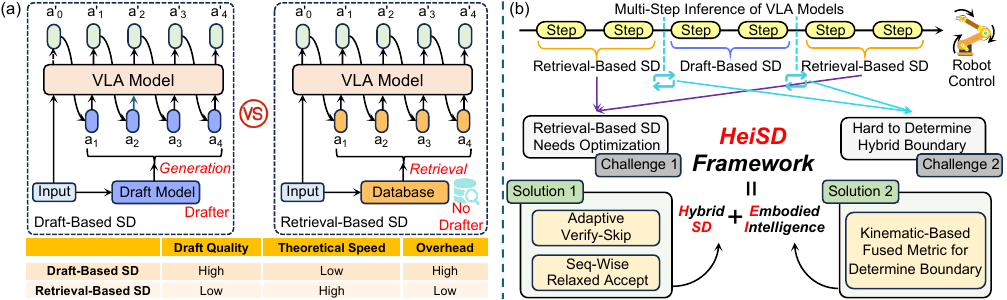}}
    \caption{Overview of the Proposed \textit{HeiSD} Framework}
    \Description{Fig.~(a) illustrates the implementation approaches of Drafter-based SD and Retrieval-based SD, while Fig.~(b) depicts the proposed \textit{HeiSD} framework.}
    \label{fig:1}
  \end{center}
  \vspace{-6mm}
\end{figure*}

% =======================================================================
% Paragraph 3: 两种SD方法各自的优势劣势是什么
% =======================================================================
When being adapted for VLA models, both drafter-based and retrieval-based SD approaches exhibit distinct advantages and limitations.
As Fig.~\ref{fig:1}~(a) shows, drafter-based SD can provide high-quality drafts with a high probability of passing verification.
But it needs to bear the inference overhead of the draft model.
Retrieval-based SD eliminates the overhead of the draft model and has a higher theoretical speedup, but suffers from low draft quality.
\textbf{This leads us to hypothesize that \underline{hybrid} using these two SD methods in VLA models may enable performance complementarity and achieve a better speed-accuracy balance.}

% =======================================================================
% Paragraph 4: 通过分析提出了关键见解，想实现混合SD是可行的/有希望达到更好性能的
% =======================================================================
To validate this hypothesis, in this study, we first construct a vector database and develop an analysis of retrieval drafts.
Our analysis reveals that some trajectory segments guided by retrieval drafts highly align with VLA inference, while others exhibit deviations. 
Based on this observation, we derive a key insight: employing retrieval-based SD for the overlapping segments and drafter-based SD for the non-overlapping segments enables leveraging the advantages of both SD approaches simultaneously.
This insight provides empirical support for our subsequent design.

% =======================================================================
% Paragraph 5: 即使理论成立，想要实现混合SD仍然面临困难
% =======================================================================
However, even with the above insight understood, achieving hybrid SD while ensuring excellent acceleration performance and minimal accuracy loss remains non-trivial, facing below challenges:
\textbf{Challenge \ding{172} :} retrieval-based SD produces low-quality drafts that fail verification easily and induce persistent errors, requiring specific optimization.
\textbf{Challenge \ding{173} :} as a posteriori metric, trajectory segments need to be converted into a priori metrics to determine the hybrid boundary, i.e., to identify which step should adopt drafter-based SD and which should use retrieval-based SD.

% =======================================================================
% Paragraph 6: 本文提出了HeiSD框架
% ======================================================================= 
In this study, we propose an end-to-end hybrid SD framework for VLA models, called \textit{HeiSD}. 
To address \textbf{Challenge \ding{172}}, \textit{HeiSD} utilizes an adaptive mechanism that selectively skips the verification process for certain drafts to avoid strict rejection.
Moreover, \textit{HeiSD} uses a sequence-wise relaxed acceptance strategy to enhance the diversity of drafts during verification and accept drafts with minor bias without compromising accuracy, to avoid persistent errors.
To address \textbf{Challenge \ding{173}}, we develop a kinematic-based fused metric in \textit{HeiSD} framework to automatically determine the hybrid boundary, enabling automatic decision and automatic switching during VLA multi-step inference. 

In summary, our contributions are three-fold:

% =======================================================================
% Paragraph 7: 本文提出了什么方法
% ========================================================================
\begin{itemize}[leftmargin=*]
    \item[$\bullet$] 
    We conducted a detailed analysis and gained key insights: hybrid using drafter-based SD and retrieval-based SD in VLA models leads to better performance.
    \item[$\bullet$] 
    We successfully optimize retrieval-based SD and propose an adaptive verify-skip mechanism along with a sequence-wise relaxed acceptance strategy, providing a basis for hybrid utilization.
    \item[$\bullet$] 
    We developed a kinematic-based fused metric to automatically determine the hybrid boundary, thus forming the \textit{HeiSD} framework. We believe \textit{HeiSD} will play a role in the future development of embodied intelligence and community building.
\end{itemize}

% =======================================================================
% Paragraph 7: 实验结果
%=======================================================================
Experimental results demonstrate that \textit{HeiSD} attains a speedup of up to 2.45$\times$ in simulation benchmarks and 2.06$\times$$\sim$2.41$\times$ in real-world scenarios, while sustaining a high task success rate.

%% file: _tex/2_background.tex
\section{Preliminary}
\label{tex:preliminary}

% =======================================================================
% subsection 1: VLA models
% =======================================================================
\subsection{Vision-Language-Action Models}
\label{tex:vla_models}
VLA models usually comprise three core components~\cite{openvla, rt2}: vision encoders (converting visual modality into tokens), an LLM backbone (fusing multi-modal information and enabling reasoning), and an action de-tokenizer (decoding output tokens into actions).
\begin{equation}
a_{j} = \mathop{\arg\max}\limits_{a_{j}}  \big [ P(a_{j} \ | \ a_{0:j-1}, \mathbb{O}, \mathbb{P}, \mathbb{W}) \big ].
\label{eq:2-1}
\end{equation}
Each VLA generation outputs an action slice, which is a 7-dimensional vector representing 7 Degrees of Freedom (DoF): position $X,Y,Z$ of the end gripper, joint rotation angles $r_{X}, r_{Y}, r_{Z}$, and a binary gripper control signal $G$.
Each DoF is encoded as a token $a_{j}$.
VLA models autoregressively predict the most probable token $a_{j}$ based on the previously tokens $a_{0:j-1}$, visual observations $\mathbb{O}$, language prompts $\mathbb{P}$, and the learnable model parameters $\mathbb{W}$, as Eq.~\eqref{eq:2-1} shown.

% =======================================================================
% subsection 2: SD
% =======================================================================
\subsection{Speculative Decoding}
\label{tex:hsd_background}
The core idea of SD is to use low-cost methods for rapid token sequence acquisition and employ LLMs for parallel verification of these tokens, thus avoiding slow autoregressive generation. 
Existing SD falls into two categories: drafter-based and retrieval-based. 
The former leverages a small draft model $M_{D}$ to generate token sequences, and uses the LLM as a verification model $M_{V}$. 
This process can be written as Eq.~\eqref{eq:2-2}, in which $a_{j}$ means the tokens, $f_{t}$ means the hidden features, and $e_{t}$ means the token embeddings.
\begin{equation}
\label{eq:2-2}
\begin{aligned}
& \textnormal{Draft:} \ a_{j} = M_{\textnormal{D}}(f_{1:t}, e_{0:t}, a_{t+1:j-1}). \\
\textnormal{Verify:} \ & \hat{a}_{j} = M_{\textnormal{V}}(a_{0:j-1}, \mathbb{P}, \mathbb{W}), \ 
\begin{cases}
\ \hat{a}_{j} = a_{j}, \ \textnormal{Accept.} \\
\ \hat{a}_{j} \neq a_{j}, \ \textnormal{Discard}.
\end{cases} 
\end{aligned}
\end{equation}
%%%
\begin{figure}[!t]
  \begin{center}
    \centerline{\includegraphics[width=3.3in]{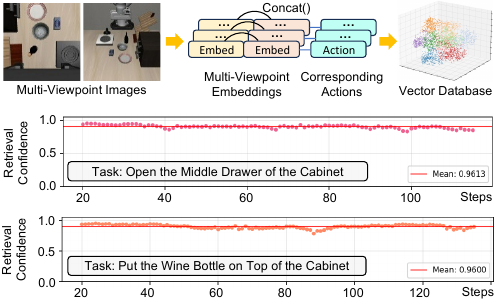}}
    \caption{Database Building Process and Corresponding Tests}
    \Description{This figure illustrates the establishment process of our database and the testing for its correctness.}
    \label{fig:add}
  \end{center}
  \vspace{-8mm}
\end{figure}
%%%%
In contrast, retrieval-based SD does not need a draft model; instead, it uses a database to retrieve draft token sequences. 
Its draft process can be written as Eq.~\eqref{eq:2-3}, where $\mathbf{DB}$ means the prebuilt database.
When applied to VLA models, the two types of SD each have distinct advantages and drawbacks. 
Drafter-based SD necessitates online maintenance of a draft model: while smaller than a VLA model, this drafter still incurs memory consumption and additional computational overhead/latency, though its high-quality drafts support longer acceptable sequences and achieve actual acceleration. 
In contrast, retrieval-based SD eliminates the overhead associated with the drafter model and offers theoretical performance benefits; however, the retrieved drafts suffer from distribution mismatch, which hinders verification and restricts the theoretical acceleration.
\begin{equation}
\label{eq:2-3}
\begin{aligned}
& \textnormal{Draft:} \ a_{j} = \big \{  Retri(f_{1:t}, e_{0:t}, a_{t+1:j-1}) \ \big | \in \mathbf{DB} \big \}. \\
& \textnormal{Verify:} \ \hat{a}_{j} = M_{\textnormal{V}}(a_{0:j-1}, \mathbb{P}, \mathbb{W}), \ 
\begin{cases}
\ \hat{a}_{j} = a_{j}, \ \textnormal{Accept.} \\
\ \hat{a}_{j} \neq a_{j}, \ \textnormal{Discard}.
\end{cases} 
\end{aligned}
\end{equation}

\begin{figure*}[!t]
  \begin{center}
    \centerline{\includegraphics[width=7in]{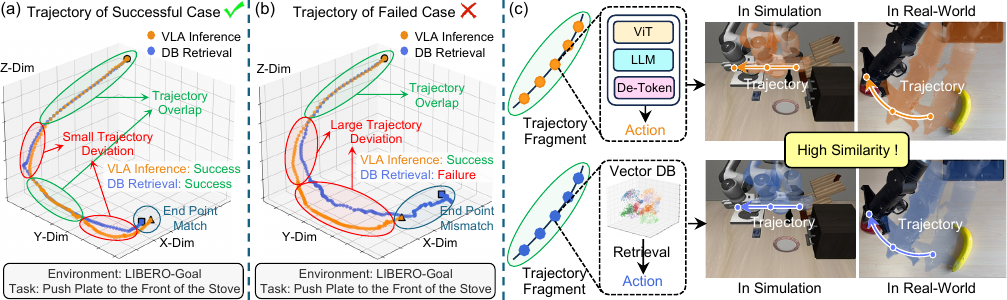}}
    \caption{Trajectory Analysis and Real-world/Simulation Validation for both VLA Inference and Database Retrieval}
    \Description{xxx}
    % 核心思路，先分析，后验证
    \label{fig:2}
  \end{center}
  \vspace{-6mm}
\end{figure*}

%% file: _tex/3_analysis.tex
\section{Observation and Motivation}
\label{tex:analysis}

This section first details the construction of the database and its performance test, then presents a database retrieval trajectory analysis. 
Based on analysis results, we derive insights into achieving hybrid SD in VLA inference and outline the corresponding challenges.

% =================================================================
% subsection 1: Database Performance Boundary
% =================================================================
\subsection{Database Building and Performance Tests}
% 数据库怎么建立的
Based on the common LIBERO datasets~\cite{LIBERO}, we first construct a robust vector database. 
As shown in the upper part of Fig.~\ref{fig:add}, we vectorize and concatenate perceptual images from multiple viewpoints in the demonstration dataset.
These images' embeddings serve as the core content stored in the database, while the corresponding demonstration actions for the current images are also stored. 
To validate the database’s generalizability, we build it with details consistent with RT-Cache~\cite{rtcache} and introduce no additional modifications.
We reuse key components including base infrastructure (based on Qdrant~\cite{qdrant}), data structure, and sharding strategy.
To maximize retrieval accuracy, all data from the LIBERO dataset are incorporated into the database.
Basic tests are conducted to test the correctness of the established database. 
In the lower part of Fig.~\ref{fig:add}, we show the retrieval confidence (Top-1) during using this built database.
Notably, the retrieval confidence of the Database is extremely high, with an average value of approximately 0.96, which validates the correctness of our database construction.
\input{tab/tab-3-1.tex}

% 单纯使用数据库有什么效果
After constructing the database and conducting correctness testing, we further explore the application of the database in embodied tasks, which serves as a precursor to retrieval-based SD.
We only use database retrieval to finish various embodied tasks.
For comparison, we report the Success Rate (SR) and speed of the OpenVLA model~\cite{openvla} inference.
As shown in Tab.~\ref{tab:3-1}, models relying solely on database retrieval can still complete certain tasks (e.g., 68.0\% in simple task suite like LIBERO-Goal) with high speed (3.74$\times$$\sim$4.83$\times$).
Moreover, the database remains effective even on challenging task suites (e.g., LIBERO-Spatial and LIBERO-Long).
This demonstrates that the constructed database holds practical application potential for embodied tasks and is highly suitable as a foundation for retrieval-based SD. 
\input{tab/tab-3-3.tex}

% =================================================================
% subsection 2: Database Retrieval Trajectory Analysis
% =================================================================
\subsection{Database Retrieval Trajectory Analysis}
% 分析了什么（轨迹）
Notably, in Tab.~\ref{tab:3-1}, the task completion rate of database retrieval is lower than that of VLA model inference.
To further investigate the causes of SR differences induced by database retrieval results, we further compare and analyze the trajectories comes from database retrieval and VLA inference.
Specifically, we use the position of the end effector of the manipulator as the measurement point and the smooth displacement between points as the trajectory.

% 分析描述
Fig.~\ref{fig:2}~(a) illustrates the trajectory where both database retrieval and VLA inference achieve task success.
It can be observed that most trajectory segments from the database and those derived from VLA inference overlap significantly (green region), which confirms the precision of the retrieved results in these segments.
Though some segments occur biased with the trajectory of VLA inference (red region), since most segments are accurate, the final endpoints can be matched (i.e. task success).
Fig.~\ref{fig:2}~(b) illustrates the trajectory where database retrieval fails but VLA inference achieves task success.
Compared with the trajectory in Fig.~\ref{fig:2}~(a), the overlapping trajectory segments are significantly reduced while the deviated ones increase substantially, leading to the failure of endpoint matching and thus the task failure.

% 得到什么结论（hybrid）
Trajectory analysis reveals that database usage should be conducted at a relatively fine granularity. 
Database retrieval maintains high accuracy for action segments corresponding to overlapping trajectories, yet may cause task failure for those associated with non-overlapping trajectories.
Therefore, we argue that database retrieval should be employed for overlapping trajectory segments, while VLA inference is suitable for non-overlapping parts.
To prove this, we develop a case study about database retrieval for overlapping segments in both simulated and real-world environments.
As shown in Fig.~\ref{fig:2}~(c), two configurations are tested: one uses VLA reasoning exclusively, and the other replaces trajectory overlapping segments with database retrieval results.
Database retrieval results for overlapping segments are highly similar to those from VLA inference in start points, end points, and motion trajectories, but with higher speed.
Guided by these, we assume that using retrieval-based SD in overlapping regions (based on the constructed database) and drafter-based SD in non-overlapping regions preserves accuracy while further improving speed, thus achieving a new Pareto Frontier.
This necessitates a hybrid SD scheme for the inference optimization of VLA models.

% =================================================================
% subsection 3: Challenges of Achieving Hybrid SD
% =================================================================
\subsection{Challenges of Achieving Hybrid SD}
However, achieving hybrid SD in VLA inference is non-trivial, facing several challenges.
% 第一个问题，Retrieval需要优化
\textbf{Challenge \ding{172}:} Achieving theoretical acceleration with retrieval-based SD for VLA models is hard.
We develop a comparison between drafter-based SD and retrieval-based SD to show the reasons, as shown in Tab.~\ref{tab:3-3}.
The distribution of retrieved drafts mismatches that of VLA inference results; thus, even if the trajectories overlap, retrieved drafts rarely pass strict verification, leading to low accept length (AL) and poor speedup.
Furthermore, due to the absence of a drafter, the system repeatedly retrieves identical results from the database, leading to persistent rejects.

% 第二个问题，很难在推理过程中自动化确定混合的边界
\textbf{Challenge \ding{173}:} Even if retrieval-based SD is effectively optimized, automatically determining the boundary of hybrid SD remains challenging.
While judging the boundary of hybrid SD by whether trajectory segments overlap is a highly effective approach, trajectory itself is a posteriori experience. 
Thus, how to predefine segments during motion becomes a key issue.

%% file: tab/tab-3-1.tex
\begin{table}[!b]
\vspace{-1mm}
\centering
\footnotesize
\caption{Database Performance on LIBERO Benchmark}
\begin{tabular}{c|c|c|c|c|c}
\toprule
\toprule
\multirow{2}{*}{\textbf{Model}} & \multirow{2}{*}{\textbf{Environment}} & \multicolumn{2}{c|}{\textbf{VLA Inference}} & \multicolumn{2}{c}{\textbf{Database Retrieval}} \\
\cmidrule{3-6}
~ & ~ & \textbf{SR} & \textbf{Speed} & \textbf{SR} & \textbf{Speed} \\
\midrule
\multirow{4}{*}{OpenVLA} & LIBERO-Goal & 77.0\% & 1.00$\times$ & 62.0\% & 4.17$\times$ \\
~ & LIBERO-Object  & 71.2\% & 1.00$\times$ & 68.0\% & 4.83$\times$ \\
~ & LIBERO-Spatial & 82.8\% & 1.00$\times$ & 53.0\% & 3.98$\times$ \\
~ & LIBERO-Long    & 54.4\% & 1.00$\times$ & 18.0\% & 3.74$\times$ \\
\bottomrule
\bottomrule
\end{tabular}
\label{tab:3-1}
\end{table}

%% file: tab/tab-3-3.tex
\begin{table}[!b]
\vspace{-1mm}
\centering
\footnotesize
\caption{Performance Comparison of Two Types of SD}
\label{tab:3-3}
\begin{tabular}{c|c|c|c|c|c|c}
\toprule
\toprule
\multirow{2}{*}{\textbf{Dataset}} & \multicolumn{3}{c|}{\textbf{VLA+SD (SpecVLA)}} & \multicolumn{3}{c}{\textbf{Retrieval-Based SD}} \\
\cmidrule{2-7}
~ & \textbf{SR} & \textbf{AL} & \textbf{Speed} & \textbf{SR} & \textbf{AL} & \textbf{Speed} \\
\midrule  
LIBERO-Goal   & 71.0\% & 2.97 & 1.00$\times$ & 77.0\% & 1.03 & 0.94$\times$ \\
LIBERO-Object & 62.4\% & 3.25 & 1.00$\times$ & 76.0\% & 0.93 & 0.92$\times$ \\
LIBERO-Spatial& 80.4\% & 3.27 & 1.00$\times$ & 78.0\% & 0.81 & 0.78$\times$ \\
LIBERO-Long   & 46.2\% & 2.82 & 1.00$\times$ & 50.2\% & 0.82 & 0.97$\times$ \\
\bottomrule
\bottomrule
\end{tabular}
\end{table}

%% file: _tex/4_rsd-optimization.tex
\section{Retrieval-Based SD Optimization}
To address the \textbf{Challenge \ding{172}} of achieving Hybrid SD in VLA Models, in this section, we propose a novel verify-skip mechanism and a seq-wise relax acceptance strategy, aiming to boost retrieval-based SD to its theoretical speed.

\begin{figure}[!t]
  \begin{center}
    \centerline{\includegraphics[width=3.3in]{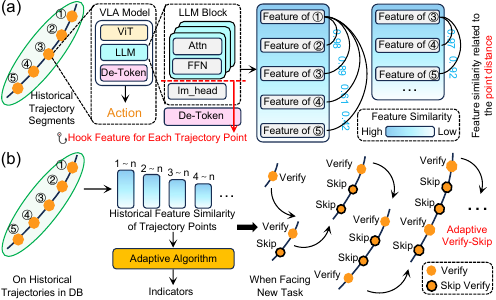}}
    \caption{Adaptive Verify-Skip Mechanism}
    \Description{xxx}
    \label{fig:3}
  \end{center}
  \vspace{-6mm}
\end{figure}

% =================================================================
% subsection 1: Adaptive Verify-Skip Mechanism
% =================================================================
\subsection{Adaptive Verify-Skip Mechanism}
\input{alg/alg-1.tex}
% 问题定义
Existing retrieval-based SD employs strict verification, which makes retrieved drafts difficult to pass the verification process and thus results in poor speedup.
The core cause of this issue lies in the fact that the draft retrieved from the database and the output results of the VLA are not strictly identically distributed in theory.
However, for multi-solution problems, the fact that the retrieved drafts and VLA's output results do not follow the same distribution does not imply incorrectness.
Some studies have attempted to explore relaxing or skipping the verification process of SD in multi-solution problems~\cite{specvla, kerv, specPV, tfl-sd, APGSD}.
We aim to follow the ideas of these studies and, by integrating the characteristics of embodied tasks, identify methods to relax or skip the verification step.

Therefore, a model-free selection strategy is required in retrieval-based SD to identify which steps’ verification should be skipped. 
To achieve this, we capture the input features of the final layer (denoted as \texttt{lm\_head}) during the verification process of each retrieved draft.
Existing work~\cite{molevla} proves that the input features of the final layer in VLA models are the most critical and most relevant to downstream robotic tasks.
We analyze the relationship between trajectory points in overlapping segments and features of the final layer, as shown in Fig.~\ref{fig:3}~(a). 
The feature similarity of trajectory points correlates with their distance: closer trajectory points exhibit higher feature similarity and should theoretically be directly accepted without verification, whereas farther trajectory points show lower similarity and should be less likely to skip verification.

Inspired by these, we employ the output features of the model's last layer as a metric to identify trajectory point eligible for verification skipping, thereby implementing an adaptive verify-skip mechanism.
Fig. 3~(b) details the proposed adaptive verify-skip mechanism.
Specifically, we calculate the feature similarity for each track point based on the historical tracks in the database.
During this process, we collect the minimum acceptable similarity $min_{S}$ and its corresponding point distance $O_{dist}$, as illustrated in the offline stage of Alg.~\ref{alg:1}.

When confronting a new task, we reuse the minimum acceptable similarity and point distance from historical information to automatically determine which trajectory points in the new task are excessively similar and thus can be skipped.
Meanwhile, feedback signals are calculated based on task completion performance. 
Using these signals, the algorithm automatically updates the minimum acceptable similarity $min_{S}$ and point distance $O_{dist}$ after each task to enhance its performance in subsequent tasks, as shown in the online stage of Alg.~\ref{alg:1}.
This adaptive adjustment mechanism enables the skipping of some trajectory point verification while allowing dynamic adjustments, thereby achieving minimal accuracy loss.

% =================================================================
% subsection 2: Adaptive Verify-Skip Mechanism
% =================================================================
\subsection{Sequence-Wise Relaxed Acceptance Strategy}

Merely addressing the failure of drafts to pass verification is insufficient; retrieval-based SD also faces the problem of persistent errors.
If errors arise in the retrieved drafts, the database lacks correction capability, resulting in repeated retrieval of the same erroneous drafts.
Consequently, the VLA model necessitates repeated verification, failing to achieve acceleration effects.
To solve this problem, we propose a sequence-wise relaxed acceptance strategy.

\begin{figure}[!t]
  \begin{center}
    \centerline{\includegraphics[width=3.3in]{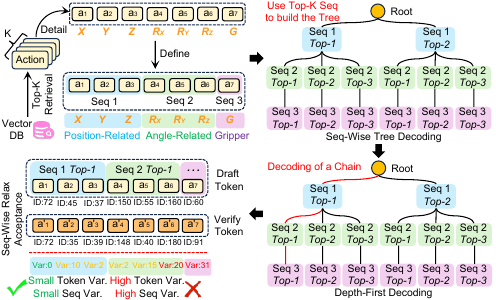}}
    \caption{Sequence-Wise Relaxed Acceptance Strategy}
    \Description{xxx}
    \label{fig:4}
  \end{center}
  \vspace{-8mm}
\end{figure}

% 改到这里了
Specifically, we first expand the retrieval draft by modifying the original Top-1 retrieval to Top-K matching. 
This approach can enhance the diversity of retrieved drafts.
Furthermore, we define sequences in the draft as a group that contains tokens with strong kinematic correlation.
As shown in Fig.~\ref{fig:4}, $X$, $Y$ and $Z$ denote the end-gripper positions (all position-related) and are thus grouped into a sequence; similarly, $R_X$, $R_Y$ and $R_Z$ represent the joint rotation angles (all angle-related) and are also packed into a sequence.
For $G$, which means the binary control signal of the gripper, 
we treat it as a separate sequence.

After that, we inherit the tree decoding backbone from Eagle-2~\cite{eagle2}.
The key difference is that we construct the tree in a sequence-wise rather than a token-wise manner.
Furthermore, when constructing the tree, we allow sequences from different retrieval results to connect with each other to maximize the potential for draft generation, except for the sequence represented by the gripper. 
This exception is justified by the strong correlation between the gripper state and task success rate.
We verify each chain starting from the root node in a depth-first manner.
In this process, a sequence-wise relaxed acceptance strategy is adopted to increase the probability of accepting diverse drafts, thereby reducing persistent errors.

As shown in Fig.~\ref{fig:4}, we calculate the index bias $bias$ of each draft token and each verify token.
When the deviation of the entire sequence $bias_{seq}$ and that of individual tokens $bias_{a_{j}}$ are maintained within a specific range, the entire sequence is accepted by enforcement.
Note that we allow $bias_{a_{j}}>\overline{bias_{seq}}$, meaning a single token may exhibit a larger bias provided that the overall bias of the sequence remains small, which differs distinctly from existing token-level relaxed acceptance.
After multiple trials, we select $bias_{seq}=30$ and $bias_{a_{j}}=15$, which achieves the optimal trade-off between accuracy and speed.
% For tree-decoding, we set the maximum nodes to 50, the tree depth to 4, and used the top 8 tokens to construct the draft tree.
After all chains are validated, if all candidates for a given sequence are rejected, the VLA model is directly activated to generate subsequent tokens. This strategy, which combines top-K retrieval with sequence-wise relaxed acceptance, effectively mitigates the issue of persistent errors.
In particular, no deviation is allowed for binary grippers, as the correctness of the gripper is critical to task success.

%% file: alg/alg-1.tex
\begin{algorithm}[!b]
  \footnotesize
  \caption{Adaptive Verify-Skip Mechanism}
  \label{alg:1}
  \begin{algorithmic}
  
% Offline Stage
    \State {\bfseries \ding{172} Offline Stage:}
    \State {\bfseries Input:} Historical Trajectory Point $P_{i}^{h}$, Number of Trajectory Point $n$, Distance $d$, Input Feature of \texttt{lm\_head} Layer $Feat_{P_{i}^{h}}$, Pre-Sampling Similarity Boundary $T$.
    \State {\bfseries Init:} $Feat_{p_{i}^{h}} \gets Hook(\texttt{lm\_head})$ after $Infer(P_{i}^{h} | P_{1 \sim i-1}^{h})$; Initialize $S(\cdot, \cdot) = 0$, $min_{S} = 0$ and $O_{dist}=0$.
    \State {\bfseries for} $i=1$ {\bfseries to} $n-1$
    \State $\quad$ {\bfseries for} $d=i$ {\bfseries to} $n-1$
    \State $\quad$ $\quad$ Compute $S(Feat_{P_{i}^{h}},Feat_{P_{i+d}^{h}})$.
    \State $\quad$ $\quad$ {\bfseries if} {$min_{S} > S(Feat_{P_{i}^{h}},Feat_{P_{i+d}^{h}}) > T$} {\bfseries then} \State $\quad$ $\quad$ $\quad$ $min_{S}, O_{dist} \gets S(Feat_{P_{i}^{h}},Feat_{P_{i+d}^{h}}), d$

    % Online Stage
    \State {\bfseries \ding{173} Online Stage:}
    \State {\bfseries Input:} Total Task Trials $t_{total}$, Trajectory Points of Current Task $P_{i}^{c}$, Number of Trajectory Points $n$, Last Task Feedback Bool Signal $B_{t}$
    \State {\bfseries Init:} Initialize $B_{n} = True$, Get $min_{S}$ and $O_{dist}$ from \ding{172}
    \State {\bfseries repeat}
    \State $\quad$ {\bfseries for} {$i=1$ {\bfseries to} $n-1$}
    \State $\quad$ $\quad$ {\bfseries for} {$d=i$ {\bfseries to} $n-1$}
    \State $\quad$ $\quad$ $\quad$ Compute $S(Feat_{P_{i}^{c}},Feat_{P_{i+d}^{c}})$.
    \State $\quad$ {\bfseries if} $B_{t}=True$ {\bfseries then} $min_{S}\textnormal{+=} \Delta|S_{c} - min(S_{h})|, O_{dist}\uparrow$
    \State $\quad$ {\bfseries else} $B_{t}=True$ {\bfseries then} $min_{S}\textnormal{-=} \Delta|S_{c} - min(S_{h})|, O_{dist}\downarrow$
    \State $t\textnormal{++}$ {\bfseries until} $t=t_{total}$
  \end{algorithmic}
\end{algorithm}

%% file: _tex/5_hybrid.tex
\section{Hybrid Boundary Determination}
\label{tex:Methods}

After optimizing retrieval-based SD, this section proposes a novel kinematic-based metric to achieve hybrid SD and automatically determine its boundary during step-by-step VLA model inference.
% =================================================================
% subsection 1: Issue Definition and Trajectory Characteristic
% =================================================================
\subsection{Issue Definition and Design Insights}

\begin{figure}[!t]
  \begin{center}
    \centerline{\includegraphics[width=3.3in]{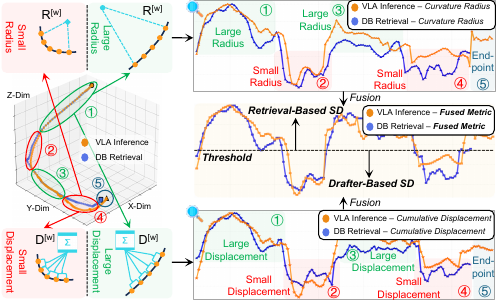}}
    \caption{Kinematic-Based Metric for Achieving Hybrid SD}
    \Description{xxx}
    \label{fig:5}
  \end{center}
  \vspace{-7mm}
\end{figure}

Aforementioned analysis in Section~\ref{tex:analysis} shows that trajectories from successful database retrievals exhibit greater overlap with VLA model inference trajectories, whereas those from failed retrievals show less overlap.
However, trajectory points are generated incrementally, and the complete trajectory cannot be obtained during the multi-step inference process.
Therefore, a method is required to determine the hybrid boundary, i.e., which steps should adopt retrieval-based SD and which should use drafter-based SD.
Through in-depth analysis, we identify that these trajectory segments exhibit inherent kinematic characteristics (shown in Fig.~\ref{fig:5}):
Overlapping trajectory segments (green region) exhibit a larger curvature radius and cumulative displacement, whereas biased segments (red region) show a smaller one.
Therefore, we decided to use these two properties to design a metric to help determine boundary.

% =================================================================
% subsection 2: Kinematic Characteristics of Trajectory Segments
% =================================================================
\subsection{Kinematic Property of Trajectory Segments}
Assuming that the trajectory context sliding window is $w$, each trajectory point in the coordinate system $\mathcal{T}(\lambda;O_\gamma)$ contains spatial position $x, y, z$, where $\lambda$ represents the distance scale and $O_\gamma$ represents the origin point.
First, we calculate the geometric center $ \mathcal{C} = (u_c,v_c) $ of trajectory segments and project it into 2-dimensional space to obtain the basis vector $(u^{[w]}_{i},v^{[w]}_{i})$, shown in Eq.~\eqref{eq:4-1}.
We use Eq.~\eqref{eq:4-2} to iteratively update the optimal geometric center, which $\mathop{Euclid}\limits_{\textnormal{2-dim}}(\cdot;\cdot)$ means 2-dimensional euclidean distance, and $\mu$ represents the mean.
After that, we use Eq.~\eqref{eq:4-3} to calculate the curvature radius $\mathcal{R}_{i}^{[w]}$.
In this way, we can scan the change of $\mathcal{R}_{i}^{[w]}$ in a sliding window $w$.
\begin{equation}
(u^{[w]}_{i},v^{[w]}_{i}) = \mathop{Proj} \big (P^{i}_{x,y,z} - \frac{1}{w}\sum_{i}^{w-1} P^{i}_{x,y,z} \ \big | \in \mathcal{T} \big ).
\label{eq:4-1}
\end{equation}
\begin{equation}
(\hat{u}_{c}, \hat{v}_{c}) = \mathop{\min}\limits_{\mathcal{C}} \sum_{i}^{w-1} \Big ( \mathop{Euclid}\limits_{\textnormal{2-dim}} \big ( (u^{[w]}_{i}, v^{[w]}_{i}); \mathcal{C} \big) - \mu \Big )^2.
\label{eq:4-2}
\end{equation}
\begin{equation}
\mathcal{R}^{[w]} = \frac{1}{w}\sum_{i}^{w-1} \mathop{Euclid}\limits_{\textnormal{2-dim}} \Big ( (u_{i}^{[w]},v_{i}^{[w]});(\hat{u}_{c},\hat{v}_{c}) \Big ).
\label{eq:4-3}
\end{equation}

Assume $\mathcal{D}^{[w]}$ means the cumulative displacement in the sliding window $w$.
We use $\mathop{Euclid}\limits_{\textnormal{3-dim}}(\cdot;\cdot)$ to calculate the displacement between two adjacent points.
Then, we summarize all the displacements to obtain $\mathcal{D}^{[w]}$ (Eq.~\ref{eq:4-4}). 
Considering that the robot sometimes performs round-trip or circular motion, $\mathcal{D}^{[w]}$ does not consider the displacement direction.
\begin{equation}
\mathcal{D}^{[w]} = \sum_{i}^{w-1} \mathop{Euclid}\limits_{\textnormal{3-dim}} \Big ( ( P^{i}_{x,y,z} \big| \in \mathcal{T} ) ; \big ( P^{i+1}_{x,y,z} \big| \in \mathcal{T} ) \Big).
\label{eq:4-4}
\end{equation}

% =================================================================
% subsection 3: Kinematic-Based Fused Metric
% =================================================================
\subsection{Kinematic-Based Fused Metric}
We fuse $\mathcal{R}^{[w]}$ and $\mathcal{D}^{[w]}$ based on Eq.~\eqref{eq:4-5} to obtain a fused metric $\mathcal{F}^{[w]}$.
$Norm(\cdot)$ means the normalization operation.
We analyze the distribution of $\mathcal{F}^{[w]}$ in supplementary materials to conclude inherent law.
Conceptually, a larger $\mathcal{F}^{[w]}$ means faster movement and a trajectory closer to a straight line (representing coarse-grained action); a smaller $\mathcal{F}^{[w]}$ results in slower movement, a more curved trajectory, and fine-grained operation.
\begin{equation}
\mathcal{F}^{[w]} = \alpha \cdot \mathop{Norm} \big (\mathcal{R}_{i}^{[w]} \big ) + (1-\alpha) \cdot \mathop{Norm}\big ( \mathcal{D}_{i}^{[w]} \big ).
\label{eq:4-5}
\end{equation}

As shown in Fig.~\ref{fig:5}, $\mathcal{F}^{[w]}$ integrates $\mathcal{R}_{i}^{[w]}$ and $\mathcal{D}_{i}^{[w]}$ to comprehensively characterize the trajectory.
We identify a threshold to serve as the demarcation between retrieval-based SD and drafter-based SD.
Specifically, when $\mathcal{F}^{[w]}$ at the current step exceeds the threshold, retrieval-based SD is selected, as both $\mathcal{R}_{i}^{[w]}$ and $\mathcal{D}_{i}^{[w]}$ at this step are relatively large, which should align with the trajectory inferred by VLA; otherwise, drafter-based SD is chosen.

% 超参数是谁？上下文窗口w和融合比例alpha

%% file: _tex/6_implementation.tex
\section{\textit{HeiSD} Framework Implementation}
\label{tex:Implementation}

This section outlines key considerations for \textit{HeiSD} framework implementation, covering cost accounting, hardware mapping and overall computation flow.

% In this section, we propose an end-to-end hybrid framework \textit{HeiSD}, solving the aforementioned problems. We will detail the design of \textit{HeiSD} below.

\begin{figure}[!t]
  \begin{center}
    \centerline{\includegraphics[width=3.3in]{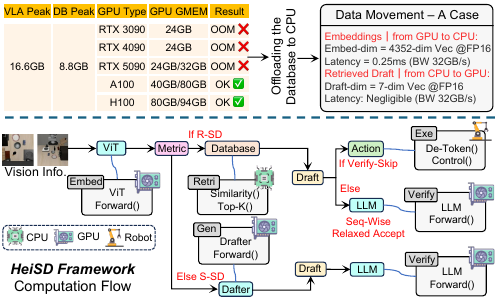}}
    \caption{System Implementation of \textit{HeiSD} Framework}
    \Description{xxx}
    \label{fig:workflow}
  \end{center}
  \vspace{-6mm}
\end{figure}

% ===============================================================
% subsection 1: Cost Accounting and Hardware Mapping
% ===============================================================
\subsection{Cost Accounting and Hardware Mapping}
% 成本核算
We develop pre-implementation cost accounting (Fig.~\ref{fig:workflow}) to reveal that most consumer GPUs encounter Out-of-Memory (OOM) issues when hosting both the database component and the VLA model; thus, to ensure compatibility with mainstream hardware, the database is deployed in CPU memory to reduce GPU memory overhead.

This heterogeneous collaborative deployment necessitates GPU-CPU communications, whose detailed costing (Fig.~\ref{fig:workflow}) shows negligible overhead ($>$100 ms) in a single inference process, far below that of model inference, validating the feasibility and effectiveness of this deployment strategy. 
Note that all GPU-related latency results are based on the NVIDIA A100. We implement the latency measurement for a single CPU-GPU data transfer using CUDA and NCU Tools.

% ===============================================================
% subsection 2: Overall Computation Flow
% ===============================================================
\subsection{Overall Computation Flow}
% 整体计算流
We use about 5000 lines of codes to implement the proposed \textit{HeiSD} framework.
The computational flow within the framework becomes complex due to the hybrid utilization of SD, thus necessitating a specialized design.
Our computation flow design is shown in Fig.~\ref{fig:workflow}.

First, visual information is encoded into embeddings using a ViT on the GPU.
This is because ViT involve substantial computational complexity, making them well-suited for efficient parallel processing on GPUs.
Then, based on the prior trajectory, our fused metric determines the type of SD to be adopted. 
If the retrieval-based SD is used, the embeddings are transmitted to the CPU for retrieval, which includes similarity computation and Top-K selection.
This is because the retrieval process involves extremely low computational complexity but consumes substantial memory, making it suitable for offloading to the CPU for execution.

When drafter-based SD is employed, the embeddings remain on the GPU and are input to the drafter for draft generation. 
This is because the computational cost of the drafter is comparable to that of ViT, so we select GPUs as the computing device for the drafter.
For the retrieved drafts, they are first moved back to the GPU.
Then the adaptive verify-skip mechanism determines whether a draft should skip verification.
If skipped, the draft is executed directly; otherwise, verification is performed under our sequence-wise relaxed acceptance strategy.
For drafts generated by the drafter, since they already reside on the GPU, we directly perform verification.
All verification processes occur on the GPU.
We will show that this CPU+GPU implementation offers advantages over the GPU-only deployment below.

%% file: _tex/7_experiments.tex
\section{Experiments}
\label{tex:experiments}

% ===============================================================
% subsection 1: Experiment Setup
% ===============================================================
\subsection{Setup}
We test \textit{HeiSD} based on OpenVLA~\cite{openvla} model clusters in LIBERO~\cite{LIBERO} simulation benchmark and real-world environment.
We build a single LLaMA block~\cite{llama} as a draft model to build drafter-based SD.
We train the draft models based on the DeepSpeed~\cite{deepspeed} framework, which takes 8 hours with 2*NVIDIA A100 GPUs.
We build retrieval-based SD using the pre-built database.
Both types of SD are applicable to OpenVLA as the validation model.
Since there is no similar work, we choose pure drafter-based SD (Pure D-SD), pure retrieval-based SD (Pure R-SD), and SOTA work SpecVLA~\cite{specvla} with token-level relaxed acceptance as the baselines.
We use an Nvidia A100 GPU and an Intel Xeon Silver 4410T as hardware platform.

% ===============================================================
% subsection 2: Evaluation Results
% ===============================================================
\subsection{Evaluation Results}
\subsubsection{\textbf{Evaluation Results on Simulation Benchmark}}
\begin{figure}[!t]
  \begin{center}
    \centerline{\includegraphics[width=3.3in]{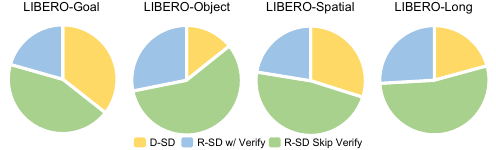}}
    \caption{Hybrid Raito and Verify-Skip Ratio in \textit{HeiSD}}
    \Description{xxx}
    \label{fig:7}
  \end{center}
  \vspace{-5mm}
\end{figure}

We utilize four LIBERO task suites to evaluate \textit{HeiSD} and each suite contains 10 tasks.
For each task, we conduct 50 trials for testing.
We report the results in Tab.~\ref{tab:6-1}.
Compared with autoregressive inference, \textit{HeiSD} achieves 1.79$\times$$\sim$2.45$\times$ speed up. 
Compared with Pure D-SD and Pure R-SD, \textit{HeiSD} also delivers significant acceleration effects.
This confirms that our choice to hybridize two types of SD during VLA reasoning is highly effective.
Moreover, even compared with SOTA works like SpecVLA, \textit{HeiSD} achieves 1.51$\times$$\sim$2.22$\times$ speedup with better SR.
This demonstrates that the hybrid use of two SD methods in the VLA reasoning process yields a superior Pareto frontier.
Fig.~\ref{fig:7} presents the proportions of two types of SD and verify-skip across four environments, explaining the source of acceleration.
\input{tab/tab-6-1.tex}
Moreover, \textit{HeiSD} achieves an increased acceptance length (AL) of about 4.75$\sim$4.96. 
The improvement in AL primarily stems from our optimization of retrieval-based SD. 
First, our verify-skip mechanism effectively extends AL, as drafts with skipped verification are treated as fully accepted. 
Second, our proposed sequence-wise relaxed acceptance strategy allows some biased tokens to be accepted alongside the sequence.

\subsubsection{\textbf{Evaluation Results on Real-World Tasks}}
\begin{figure*}[!t]
  \begin{center}
    \centerline{\includegraphics[width=6.7in]{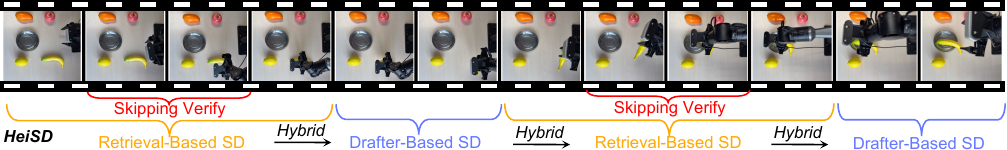}}
    \caption{A Case of \textit{HeiSD} Framework Completing Real-World Tasks (Task Name: Pick up the Banana and Put It on the Plate)}
    \Description{xxx}
    \label{fig:8}
  \end{center}
  \vspace{-5mm}
\end{figure*}

We construct a tabletop operating environment, and employ the AgileX PIPER robot arm to test \textit{HeiSD}'s performance in real-world scenarios.
We build various manipulation tasks.
We collect massive human demonstration data to rebuild the database and fine-tune the models.
Details regarding the tabletop operating environment, robotic arm specifications, task design, and model fine-tuning are provided in detail in the supplementary materials.
After that, we deploy \textit{HeiSD} into the real world and test its performance.
The results are shown in Tab.~\ref{tab:6-2}.
On the three task categories we defined (with SR of 87.2\%, 77.3\%, and 71.7\% after fine-tuning), \textit{HeiSD} achieves a 2.06$\times$$\sim$2.41$\times$ speedup with minor SR loss (1.2\%$\sim$3.9\%).
We also represent an example real-world task completion process in Fig.~\ref{fig:8} to show the performance.

\input{tab/tab-6-2.tex}

\subsubsection{\textbf{Ablation Studies}}
We conduct ablation studies on the LIBERO-Goal benchmark to evaluate the effects of \textit{HeiSD}'s components, with corresponding results reported in Tab.~\ref{tab:6-2}.
When Hybrid SD is implemented solely based on fusion metrics, the boundaries between retrieval-based SD and drafter-based SD are defined; however, the overall performance remains suboptimal due to unresolved issues in retrieval-based SD.
Specifically, it achieves an SR of 74.0\% but exhibits nearly no acceleration, with an average accepted length of only 1.05.
Building on this foundation, we incorporate the adaptive verify-skip mechanism, which significantly enhances performance.
It achieves a 2.08$\times$ speedup with only a slight drop (1.0\%) in accuracy, while the AL is also increased to 4.04.
Further, after adding the sequence-wise relaxed acceptance, it achieves 2.38$\times$ speedup (0.3$\times$ increasing) and an AL of 4.50, while maintaining 73.0\% SR.

\input{tab/tab-6-3.tex}

% ===============================================================
% subsection 3: Discussion
% ===============================================================
\subsection{Discussion}
\begin{figure}[!t]
  \begin{center}
    \centerline{\includegraphics[width=3.3in]{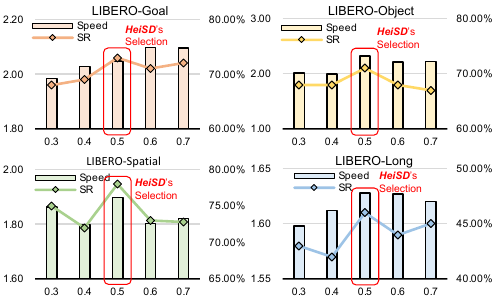}}
    \caption{Discussion of Hyper-Parameter $w$ in \textit{HeiSD}}
    \Description{xxx}
    \label{fig:10}
  \end{center}
  \vspace{-6mm}
\end{figure}

\begin{figure}[!t]
  \begin{center}
    \centerline{\includegraphics[width=3.3in]{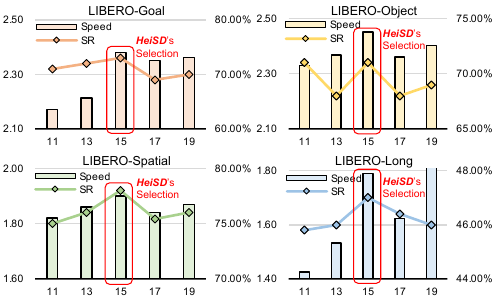}}
    \caption{Discussion of Hyper-Parameter $\alpha$ in \textit{HeiSD}}
    \Description{xxx}
    \label{fig:9}
  \end{center}
  \vspace{-6mm}
\end{figure}

\noindent \textbf{Hyper-Parameters.}
\textit{HeiSD} involves two hyper-parameters: the sliding window size $w$ for trajectory points and $\alpha$ in the fused metric.
We evaluate \textit{HeiSD}'s performance across various hyper-parameters on the LIBERO-Goal benchmark, with the results presented in Fig.~\ref{fig:9}.
As illustrated in Fig.~\ref{fig:9}, we test the results under different values of $w$ across four simulation environments. 
We find that the value of $w$ exerts a significant influence on speed and SR, as it alters the distribution of the fused metric.
Considering both the SR and speed comprehensively, we select $w = 15$ as the default value for \textit{HeiSD}.
Fig.~\ref{fig:10} shows the results under different values of $\alpha$.
Similarly, $\alpha$ also influences the SR and speed. 
We select $\alpha=0.5$ to ensure equal attention is allocated to the curvature radius and cumulative displacement.

% 超参数1: 滑动窗口w的大小
% 超参数2: 融合指标的比例\alpha

\noindent \textbf{Hardware Implementation Analysis.}
We test \textit{HeiSD}'s end-to-end latency (50 trials) on different hardware platforms in both four simulation benchmarks.
Results in Tab.~\ref{tab:6-4} show that offload database into CPU brings a little acceleration (1.04$\times$$\sim$1.09$\times$).
This demonstrates that the CPU exhibits higher execution efficiency than the GPU for database retrieval-related operations without specific optimization.
Notably, the core reason for selecting CPU+GPU collaborative deployment is to reduce the GPU memory footprint of the database in Retrieval-based Stable Diffusion (Retrieval-based SD).
Specifically, quantifying the speedup achieved by the introduction of the CPU is not our core objective.
\input{tab/tab-6-4.tex}

\noindent \textbf{Generality.}
The proposed HeiSD framework exhibits broad applicability.
This is because, despite the structural diversity of existing VLA models, autoregressive-based LLMs remain their core component. 
Even if the VLA model introduces diffusion-based action generators, it still requires LLM to generate intermediate features autoregressively. 
In this case, our design can continue to play an accelerating role.
Moreover, the proposed \textit{HeiSD} framework exhibits good generality and imposes no specific requirements on task categories or robot platforms.
Our design is based on the kinematic characteristics of trajectories, which rely on physical principles and are independent of both the task and the robotic arm platform.

\noindent \textbf{Scope.}
We do not design an automatic hyperparameter determination method, mainly because established standards for current embodied intelligence are lacking. 
Automatic hyper-parameter selection is therefore designated as future work, to be addressed once environmental standards are established. 
Additionally, the core purpose of \textit{HeiSD} is to achieve acceleration while ensuring SR, rather than to guarantee strictly homogeneous distribution of draft and inference results.
Thus, the impact of verify-skip and sequence-wise relaxed acceptance on the output distribution is not considered in this paper as it falls outside the scope.
This aligns with the state-of-the-art (SOTA) works~\cite{specPV, APGSD, tfl-sd}.

%% file: tab/tab-6-1.tex
\begin{table}[!b]
    \centering
    \setlength{\tabcolsep}{1.4mm}
    \caption{Simulation Results of \textit{HeiSD}}
    \label{tab:6-1}
    \footnotesize
    \begin{tabular}{c|c|c|c|c|c|c}
    \toprule
    \toprule
    {\textbf{Env.}} & \textbf{Method} & {\textbf{SR}} & \textbf{Speed} & \textbf{AL} & \textbf{Steps} & \textbf{HW} \\
    \midrule
    \multirow{5}{*}{LIBERO-Goal} & \cellcolor{gray!20}{AR w/o SD} & \cellcolor{gray!20}{77.0\%} & \cellcolor{gray!20}{1.00$\times$} & \cellcolor{gray!20}{--} & \cellcolor{gray!20}{157.6} & \cellcolor{gray!20}{GPU} \\
    ~ & \cellcolor{pink!20}{Pure R-SD} & \cellcolor{pink!20}{77.0\%} & \cellcolor{pink!20}{0.96$\times$} & \cellcolor{pink!20}{1.03} & \cellcolor{pink!20}{152.6} & \cellcolor{pink!20}{CPU+GPU} \\
    ~ & \cellcolor{orange!20}{Pure D-SD} & \cellcolor{orange!20}{76.2\%} & \cellcolor{orange!20}{0.87$\times$} & \cellcolor{orange!20}{1.68} & \cellcolor{orange!20}{159.2} & \cellcolor{orange!20}{GPU} \\
    ~ & \cellcolor{yellow!20}{SpecVLA} & \cellcolor{yellow!20}{71.0\%} & \cellcolor{yellow!20}{1.23$\times$} & \cellcolor{yellow!20}{3.63} & \cellcolor{yellow!20}{166.8} & \cellcolor{yellow!20}{GPU} \\
    ~ & \cellcolor{green!20}{\textbf{\textit{HeiSD}}} & \cellcolor{green!20}{\textbf{73.0\%}} & \cellcolor{green!20}{\textbf{2.38$\times$}} & \cellcolor{green!20}{\textbf{4.75}} & \cellcolor{green!20}{\textbf{156.3}} & \cellcolor{green!20}{\textbf{CPU+GPU}} \\
    \cmidrule{1-7}
    \multirow{5}{*}{LIBERO-Object} & \cellcolor{gray!20}{AR w/o SD} & \cellcolor{gray!20}{71.2\%} & \cellcolor{gray!20}{1.00$\times$} & \cellcolor{gray!20}{--} & \cellcolor{gray!20}{191.7} & \cellcolor{gray!20}{GPU} \\
    ~ & \cellcolor{pink!20}{Pure R-SD} & \cellcolor{pink!20}{76.0\%} & \cellcolor{pink!20}{0.98$\times$} & \cellcolor{pink!20}{0.93} & \cellcolor{pink!20}{195.4} & \cellcolor{pink!20}{GPU} \\
    ~ & \cellcolor{orange!20}{Pure D-SD} & \cellcolor{orange!20}{68.6\%} & \cellcolor{orange!20}{0.96$\times$} & \cellcolor{orange!20}{1.84} & \cellcolor{orange!20}{195.9} & \cellcolor{orange!20}{GPU} \\
    ~ & \cellcolor{yellow!20}{SpecVLA} & \cellcolor{yellow!20}{62.4\%} & \cellcolor{yellow!20}{1.10$\times$}  & \cellcolor{yellow!20}{3.91} & \cellcolor{yellow!20}{214.5} & \cellcolor{yellow!20}{GPU}\\
    ~ & \cellcolor{green!20}{\textbf{\textit{HeiSD}}} & \cellcolor{green!20}{\textbf{71.0\%}} & \cellcolor{green!20}{\textbf{2.45$\times$}} & \cellcolor{green!20}{\textbf{4.94}} & \cellcolor{green!20}{\textbf{189.6}} & \cellcolor{green!20}{\textbf{CPU+GPU}} \\
    \cmidrule{1-7}
    \multirow{5}{*}{LIBERO-Spatial} & \cellcolor{gray!20}{AR w/o SD} & \cellcolor{gray!20}{82.8\%} & \cellcolor{gray!20}{1.00$\times$} & \cellcolor{gray!20}{--} & \cellcolor{gray!20}{126.9} & \cellcolor{gray!20}{GPU} \\
    ~ & \cellcolor{pink!20}{pure R-SD} & \cellcolor{pink!20}{78.0\%} & \cellcolor{pink!20}{1.15$\times$} & \cellcolor{pink!20}{0.81} & \cellcolor{pink!20}{123.5} & \cellcolor{pink!20}{GPU} \\
    ~ & \cellcolor{orange!20}{Pure D-SD} & \cellcolor{orange!20}{82.8\%} & \cellcolor{orange!20}{0.98$\times$} & \cellcolor{orange!20}{1.65} & \cellcolor{orange!20}{127.3} & \cellcolor{orange!20}{GPU} \\
    ~ & \cellcolor{yellow!20}{SpecVLA} & \cellcolor{yellow!20}{\textbf{80.4\%}} & \cellcolor{yellow!20}{1.26$\times$}  & \cellcolor{yellow!20}{3.80} & \cellcolor{yellow!20}{128.7} & \cellcolor{yellow!20}{GPU} \\
    ~ & \cellcolor{green!20}{\textbf{\textit{HeiSD}}} & \cellcolor{green!20}{\textbf{78.0\%}} & \cellcolor{green!20}{\textbf{1.90$\times$}} & \cellcolor{green!20}{\textbf{4.83}} & \cellcolor{green!20}{\textbf{127.8}} & \cellcolor{green!20}{\textbf{CPU+GPU}} \\
    \cmidrule{1-7}
    \multirow{5}{*}{LIBERO-Long} & \cellcolor{gray!20}{AR w/o SD} & \cellcolor{gray!20}{54.4\%} & \cellcolor{gray!20}{1.00$\times$} & \cellcolor{gray!20}{--} & \cellcolor{gray!20}{393.2} & \cellcolor{gray!20}{GPU} \\
    ~ & \cellcolor{pink!20}{Pure R-SD} & \cellcolor{pink!20}{50.0\%} & \cellcolor{pink!20}{0.99$\times$} & \cellcolor{pink!20}{0.81} & \cellcolor{pink!20}{399.3} & \cellcolor{pink!20}{GPU} \\
    ~ & \cellcolor{orange!20}{Pure D-SD} & \cellcolor{orange!20}{50.2\%} & \cellcolor{orange!20}{0.91$\times$} & \cellcolor{orange!20}{1.59} & \cellcolor{orange!20}{400.7} & \cellcolor{orange!20}{GPU} \\
    ~ & \cellcolor{yellow!20}{SpecVLA} & \cellcolor{yellow!20}{\textbf{46.2\%}} & \cellcolor{yellow!20}{1.13$\times$} & \cellcolor{yellow!20}{3.63} & \cellcolor{yellow!20}{439.6} & \cellcolor{yellow!20}{GPU} \\
    ~ & \cellcolor{green!20}{\textbf{\textit{HeiSD}}} & \cellcolor{green!20}{\textbf{47.0\%}} & \cellcolor{green!20}{\textbf{1.79$\times$}} & \cellcolor{green!20}{\textbf{4.96}} & \cellcolor{green!20}{\textbf{428.0}} & \cellcolor{green!20}{\textbf{CPU+GPU}} \\
    \bottomrule
    \bottomrule
    \end{tabular}
\end{table}

%% file: tab/tab-6-2.tex
\begin{table}[!b]
\centering
\footnotesize
\setlength{\tabcolsep}{1mm}
\caption{Real-World Results of \textit{HeiSD}}
\label{tab:6-2}
\begin{tabular}{l|c|c|c|c}
\toprule
\toprule
\textbf{Task Category} & \textbf{Fine-Tune} & \textbf{\textit{HeiSD} SR} & \textbf{ \textit{HeiSD} Speedup} & \textbf{\textit{HeiSD} AL} \\
\midrule
Atomic Grasping      & 87.2\% & 86.0\% & 2.33$\times$ & 4.47 \\
Spatial Displacement & 77.3\% & 75.1\% & 2.41$\times$ & 4.39 \\
Composite Sequential & 71.7\% & 67.8\% & 2.06$\times$ & 4.15 \\
\bottomrule
\bottomrule
\end{tabular}
\end{table}

%% file: tab/tab-6-3.tex
\begin{table}[!b]
\centering
\footnotesize
\caption{Ablation Studies of \textit{HeiSD} on LIBERO-Goal}
\label{tab:6-3}
\begin{tabular}{l|l|l|l}
\toprule
\toprule
~ & \textbf{SR} & \textbf{Speedup} & \textbf{AL} \\
\midrule
Only Hybrid SD & 74.0\% & 1.05$\times$ & 1.05 \\
+ Adaptive Verify-Skip & 73.0\% \textcolor{red}{$\downarrow$1.0\%} & 2.08$\times$ \textcolor{green}{$\uparrow$1.03$\times$} & 4.04 \textcolor{green}{$\uparrow$2.99} \\
+ Seq-Wise Relaxed Acceptance & 73.0\% \textcolor{green}{$\uparrow$0.0\%} & 2.38$\times$ \textcolor{green}{$\uparrow$0.30$\times$} & 4.50 \textcolor{green}{$\uparrow$0.46} \\
\bottomrule
\bottomrule
\end{tabular}
\end{table}

%% file: tab/tab-6-4.tex
\begin{table}[!b]
\centering
\footnotesize
\caption{Hardware Implementation Analysis of \textit{HeiSD}}
\label{tab:6-4}
\begin{tabular}{l|c|c|c|c}
\toprule
\toprule
\textbf{Details} & \textbf{Goal} & \textbf{Object} & \textbf{Spatial} & \textbf{Long} \\
\midrule
\textit{HeiSD} on GPU & 7388.9 s & 7731.7 s & 6557.0 s & 23318.5 s \\
\textit{HeiSD} on CPU+GPU & 7113.6 s & 7328.4 s & 6238.5 s & 21352.7 s \\
\cmidrule{2-5}
Acceleration & 1.04$\times$ & 1.06$\times$ & 1.05$\times$ & 1.09$\times$ \\
\bottomrule
\bottomrule
\end{tabular}
\end{table}

%% file: _tex/8_conclusion.tex
\section{Conclusion}
\label{tex:conclusion}

In this paper, we first construct a database, test its performance and correctness, identify the trajectory overlapping regularity.
We derive a key insight that implementing hybrid SD for VLA inference optimization will achieve better performance.
Based on this insight, we introduce an adaptive verify-skip mechanism and a sequence-wise relaxed acceptance strategy to address limitations in retrieval-based SD. 
Additionally, we develop a kinematic-based fusion metric to determine the hybrid SD boundary, forming the \textit{HeiSD} framework.
We conduct various experiments to evaluate the advances of the \textit{HeiSD} framework in both simulation benchmarks and real-world scenarios.
Experiments show that \textit{HeiSD} attains a speedup of up to 2.45$\times$ in simulation benchmarks and 2.06$\times$$\sim$2.41$\times$ in real-world scenarios, while sustaining a high SR.

%% file: _tex/9_appendix.tex
% ===================================================================
% section 1: VLA model
% ===================================================================
\section{Vision-Language-Action Models}
\label{apdx:VLA models}

\subsection{Model Structure}
\label{apdx:VLA models-structure}
% VLA模型的基本结构，并简单说明AM是不包含在本文范围内的。 - 已检查
Vision-Language-Action (VLA) models map visual observations and natural language instructions directly to robot actions. 
The canonical VLA architecture follows a three-stage pipeline: (1) a \textbf{Vision Encoder} (ViT-based) extracts visual features from robot observations, (2) a pre-trained \textbf{Large Language Model} backbone fuses visual and language tokens for cross-modal reasoning, and (3) an \textbf{Action Decoder} head projects the model outputs into action space. 
While recent works explore diffusion-based action generation, their core still relies on autoregressive VLMs, making our proposed \textit{HeiSD} framework applicable.

\begin{figure}[!b]
\centering
\includegraphics[width=7in]{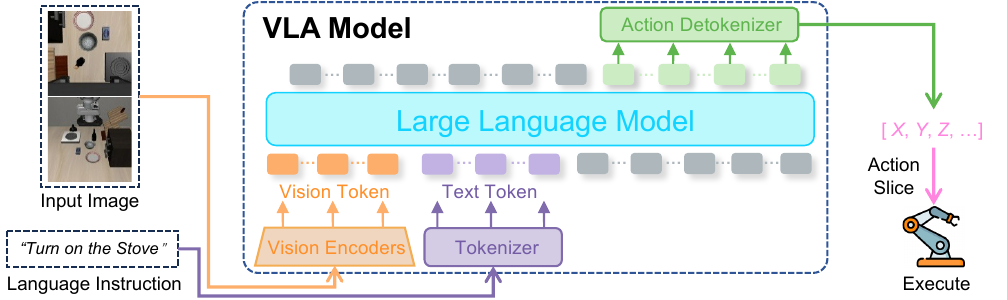}
\caption{Common Structure of Vision-Language-Action Models.}
\label{fig:vla-arch}
\end{figure}

To illustrate this architecture concretely, we use OpenVLA~\cite{openvla} as an example. 
OpenVLA employs a dual-encoder vision system: DINOv2-ViT-L/14~\cite{dinov2} (304M parameters, self-supervised features) and SigLIP-ViT-SO400M/14~\cite{siglip} (400M parameters, vision-language aligned features) process $224 \times 224$ RGB images in parallel. 
Their 1024-dim and 1152-dim outputs are concatenated and projected through a lightweight adapter into the LLM token space, yielding 256 visual tokens per observation. 
The language backbone is Llama-2-7B (7B parameters), which autoregressively generates action tokens by attending to the combined vision-language context. 
For 7-DoF robot control, the model predicts 7 action dimension tokens through a linear head over a 256-bin vocabulary per dimension.

\subsection{Generation Paradigms}
\label{apdx:VLA models-Paradigms}
At each robot control timestep, action generation proceeds sequentially: the model predicts dimension $a_1$, appends its token to the input sequence, then predicts $a_2$ conditioned on $a_1$, and so forth. 
For a $D$-dimensional action ($D=7$ for OpenVLA), this requires $D$ full forward passes through the LLM backbone, each involving causal attention over all visual tokens, language tokens, and previously generated action tokens. 
On an NVIDIA A100 GPU, OpenVLA's single-timestep inference takes approximately 174ms: 8ms for dual-encoder visual feature extraction, 113ms for autoregressive action token generation (across 7 decoding steps), and $\approx 53$ms of system overheads (e.g., data transfer and CPU scheduling).
This sequential decoding bottleneck—where each action dimension must wait for its predecessor—directly motivates our application of speculative decoding to VLA inference.

% token-action mapping是怎样的。 - 已检查，公式拆到单独行，详细描述。
To bridge continuous robot control and discrete language modeling, VLA models quantize action spaces into token vocabularies. Each action dimension $a_i \in [a_{\min,i}, a_{\max,i}]$ is uniformly discretized into $K$ bins (typically $K=256$ to balance resolution and vocabulary size). 
The continuous-to-discrete mapping assigns each action to its nearest bin index:
\begin{equation}
b_i = \left\lfloor \frac{a_i - a_{\min,i}}{a_{\max,i} - a_{\min,i}} \cdot (K-1) \right\rfloor,
\end{equation}
while the inverse mapping reconstructs continuous actions via linear interpolation:
\begin{equation}
a_i = a_{\min,i} + \frac{b_i}{K-1}(a_{\max,i} - a_{\min,i}).
\end{equation}
This discretization allows VLA models to apply standard cross-entropy loss over bin indices, treating action prediction as multi-class classification without architectural modifications.

% ===================================================================
% section 2: LIBERO Dataset Details
% ===================================================================
\section{LIBERO Dataset Details}
\label{apdx:LIBEROdataset}

\subsection{Brief Introduction of LIBERO}
\label{apdx:LIBEROdataset-intro}
LIBERO~\cite{LIBERO} is a comprehensive benchmark for evaluating multitask robot learning and generalization.
The dataset contains 130 tabletop manipulation tasks performed by a 7-DoF Franka Panda robot arm in simulated environments built on MuJoCo~\cite{mujoco} physics engine.
All demonstrations are collected through human tele-operation using expert operators, ensuring high-quality trajectory data.
Depending on the task environment and operands, LIBREO classifies all tasks into four categories, as follows:
\begin{itemize}
\item \textbf{LIBERO-Spatial}: Tasks vary in spatial configurations and object placements while maintaining consistent manipulation primitives (e.g., ``pick up the bowl on the left'' and ``pick up the bowl on the right'').
\item \textbf{LIBERO-Object}: Tasks involve different object instances with varying visual appearances but similar manipulation strategies (e.g., different colored plates, various shaped containers).
\item \textbf{LIBERO-Goal}: Tasks require the same set of objects but with different goal specifications, testing the agent's ability to follow diverse instructions.
\item \textbf{LIBERO-Long}: Multi-step tasks requiring sequential execution of 3-4 sub-goals, significantly longer than single-step manipulation tasks in other suites.
\end{itemize}

Each task provides multiple demonstration trajectories with randomized initial states, enabling robust policy learning and systematic evaluation of generalization capabilities across different distribution shifts.

\begin{figure}[!t]
  \begin{center}
    \centerline{\includegraphics[width=7in]{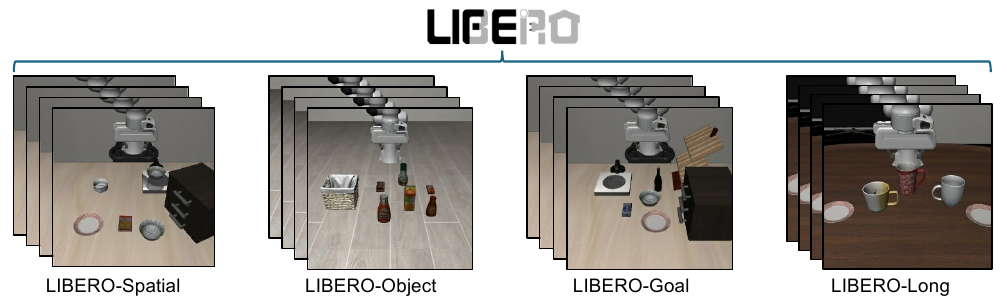}}
    \caption{Representative initial scenes from the four LIBERO subsets. \textnormal{From left to right: LIBERO-Spatial, LIBERO-Object, LIBERO-Goal, and LIBERO-Long. These MuJoCo tabletop environments, manipulated by a 7-DoF Franka Panda arm, illustrate distribution shifts across spatial configurations, object instances,and goal specifications.}}
  \end{center}
  \label{fig:LIBERO_Enviroment}
  \vskip -0.3in
\end{figure}

\input{tab/apdx-LIBEROstat}

\subsection{Statistics of the LIBERO Dataset}
\label{apdx:LIBEROdataset-statistics}

We build our retrieval database from the official LIBERO training demonstrations~\cite{LIBERO}. 
To guide the design of the database architecture, we profile key dataset statistics—such as episode count, length, Disk Usage, and action dimensionality—across its four task suites: LIBERO-Goal, LIBERO-Spatial, LIBERO-Object, and LIBERO-Long. 
As shown in Tab.~\ref{tab:dataset_stats}, LIBERO-Long episodes are significantly longer due to multi-step task composition, while the others focus on single-goal primitives.

% ===================================================================
% section 3: Details of Retrieval-Based Optimization
% ==================================================================
\section{Database Construction}
\label{apdx:Database}

This section details the design and implementation of our retrieval system, which enables fast, searching over human demonstrations. 
We first describe the multi-modal vector representation used to encode visual states, followed by the self-contained payload schema that associates each vector with executable action sequences and metadata. 
Based on scale profiling of the LIBERO dataset, we then present our database architecture—including backend selection, and memory feasibility. 
Finally, we report the actual storage footprint and query latency of the constructed database, demonstrating its efficiency as a low-latency retrieval module.

\subsection{Construction Details of the Retrieval System}
\label{apdx:Database-Construction}

\noindent \textbf{Vector Representation.} Vision-Language-Action (VLA) models often incorporate multiple visual encoders or Vision Transformer (ViT) backbones to capture complementary aspects of the environment. 
To construct a comprehensive observation representation, we fuse the visual features from these complementary streams by concatenation.
Taking OpenVLA~\cite{openvla} as a concrete example, its visual encoder provides two distinct feature types:
\begin{itemize}
    \item \textbf{DINOv2 Features} (1024-dim): A self-supervised visual representation that captures robust scene-level semantics.
    \item \textbf{SigLIP Features} (1152-dim): A vision-language aligned visual representation that grounds observations to task instructions.
\end{itemize}
In our setup, we employ a dual-view observation system consisting of a third-person view and an elbow-mounted camera view.
Each view is independently processed by both encoders, yielding $(1024 + 1152) \times 2 = 4352$ dimensions in total.

\begin{figure}[!t]
  \begin{center}
    \centerline{\includegraphics[width=7in]{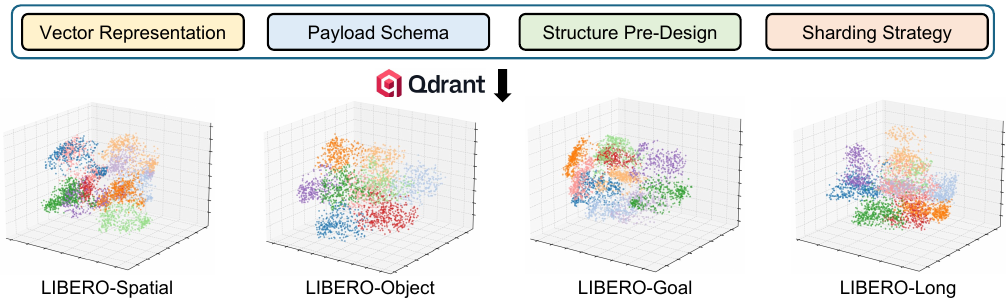}}
    \caption{3D visualization of database embedding vectors}
    \label{fig:db-visual}
  \end{center}
  \vskip -0.3 in
\end{figure}

Concatenating features from both views yields a 4352-dim joint visual embedding, which is then L2-normalized to enable cosine similarity-based retrieval. 
Following Qdrant's efficient retrieval methodology, we leverage Hierarchical Navigable Small World (HNSW) graphs for approximate nearest neighbor search, which provide sub-linear query complexity while maintaining high recall rates.

\noindent \textbf{Structure Pre-Design.} 
Guided by our profiling of the LIBERO benchmark—comprising 273,465 timesteps across four suites (summing to fewer than $3 \times 10^5$ steps.
We determine that the entire retrieval memory can comfortably fit in RAM. 
Each timestep entry requires approximately 8.69~KiB, leading to a theoretical total memory footprint of 2.27~GB.
Given these constraints, we select Qdrant~\cite{qdrant} as the vector database backend due to its high-speed similarity search, open-source licensing, and native support for loading full indices into memory. 
Qdrant employs HNSW indexing with configurable parameters to balance index construction time, memory usage, and search accuracy.
This design ensures low-latency retrieval during policy inference while maintaining simplicity in deployment.

\input{tab/apdx-DBdetails.tex}

\subsection{Performance Analysis of the Retrieval System}
\label{apdx:Database-Performance}

As summarized in Tab.~\ref{tab:db_stats}, the final database contains 273,465 vectors—one for each timestep in the LIBERO training set.
We construct it by encoding each observation into a 4352-dim vector and storing it together with its payload, ensuring complete, non-redundant coverage.

The actual disk usage is 6.5~GB, comprising 1.0~GB for vector storage and 5.5~GB for indexing overhead (HNSW graphs and metadata).
We measure retrieval latency on a standard server: the average query time over 100 runs is \textbf{5.13~ms}, significantly faster than a single SD forward pass (13.93~ms).
This confirms that our retrieval memory is both compact and efficient for real-time action drafting.

\noindent \textbf{Retrieval Confidence Analysis.}
Beyond query latency, we further analyze the reliability of retrieval results through confidence measurement. 
In our system, retrieval confidence is quantified by the cosine similarity score between the query embedding and its nearest neighbor in the database—higher scores indicate closer matches between the current observation and stored demonstrations.

\begin{figure}[!t]
  \begin{center}
    \centerline{\includegraphics[width=7in]{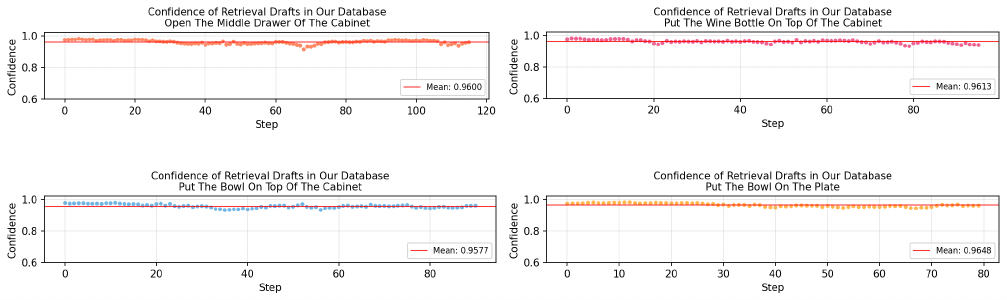}}
    \caption{Distribution of retrieval confidence (cosine similarity scores) across the four LIBERO task suites. The consistently high scores ($>$0.90 for most queries) demonstrate the reliability of our constructed database for action retrieval.}
    \label{fig:DBconfidence}
  \end{center}
  \vskip -0.4 in
\end{figure}

As illustrated in Fig.~\ref{fig:DBconfidence}, the retrieval confidence scores are consistently high across all four LIBERO task suites, with the majority of queries achieving similarity scores above 0.90.
This uniformly high confidence demonstrates that our database construction effectively captures the visual-semantic patterns of demonstration trajectories, enabling reliable nearest-neighbor matching during inference.
The consistently high retrieval confidence validates the quality of our fused visual embedding and the completeness of our database coverage over the LIBERO demonstration space.

\begin{figure}[!b]
  \begin{center}
    \vskip -0.2 in
    \centerline{\includegraphics[width=7in]{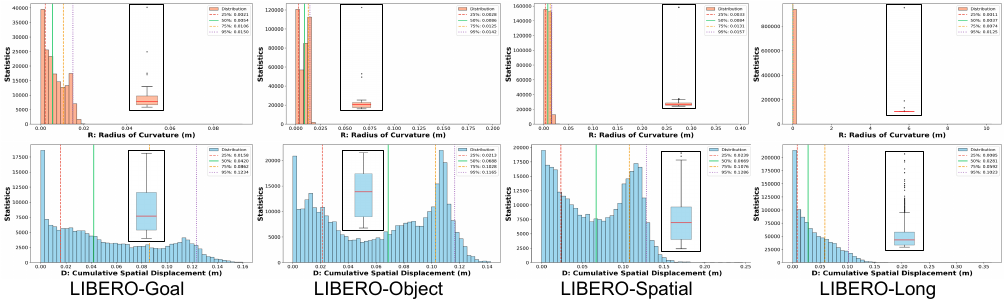}}
    \caption{Empirical distributions of Cumulative Spatial Displacement ($D^{[w]}$) and Radius of Curvature ($R^{[w]}$) across the four LIBERO task suites. The dashed lines indicate the 95$^{\textnormal{th}}$ percentile thresholds used for normalization.}
    \label{fig:RnD-distribution}
  \end{center}
\end{figure}

\subsection{Package and Application of the Retrieval System}
\label{apdx:Database-application}

\noindent \textbf{System Architecture.}
To support real-time integration with VLA policies, we implement the retrieval system as a lightweight \textit{Retrieval Class} that can be directly invoked by the policy agent. 
The retrieval class encapsulates two core components: (1) an \textit{Embedding Module} that processes dual-view input images (third-person and elbow-mounted camera) into 4352-dimensional vectors by fusing frozen DINOv2 and SigLIP features from both views via concatenation and L2 normalization, leveraging GPU acceleration for low-latency encoding; and (2) a \textit{Qdrant Database} client that maintains the vector index with HNSW graphs for efficient similarity search. 
Critically, each vector is paired with a self-contained JSON payload that embeds all necessary action and metadata—eliminating the need for external storage or cross-database queries, as required by systems like RT-Cache~\cite{rtcache}.

\noindent \textbf{Offline Construction Pipeline.}
The database is constructed in a single offline pass over the LIBERO RLDS datasets. For each timestep containing an observation, language instruction, and action sequence, we (1) generate its fused visual embedding using the same encoder configuration as at inference time, (2) construct a payload that includes metadata, the current 7-DoF action, and a three-step lookahead action sequence, and (3) insert the (embedding, payload) pair into the Qdrant database. 
After ingestion, HNSW indices are built to support fast approximate nearest neighbor search during runtime.

\noindent \textbf{Online Retrieval Workflow.}
During policy execution, the system enables low-latency speculative drafting through direct function calls. 
Given a new observation and instruction from the VLA agent, the retrieval class (1) computes a query vector using the embedded encoding module with identical frozen encoders, (2) performs a top-K approximate nearest neighbor search via HNSW, (3) retrieves the payload of the nearest neighbor, and (4) extracts the action field to return a draft 3-step action sequence.  
The full retrieval pipeline achieves an average latency of \textbf{5.13~ms} (see Appendix~\ref{apdx:Database-Performance}), well within real-time requirements for interactive robotic control.

\section{Normalization of Kinematic-Based Metric}
\label{apdx:MetricNorm}

In \textit{HeiSD}, we employ two kinematic indicators—\textbf{Cumulative Spatial Displacement} ($D^{[w]}$) and \textbf{Radius of Curvature} ($R^{[w]}$)—to assess trajectory smoothness and guide the adaptive switching mechanism. 
Since these raw metrics exhibit different scales and distributions across task suites, proper normalization is essential to ensure consistent and comparable thresholding.

\subsection{Profiling of Raw Indicator Distributions}

To characterize the natural range of each indicator, we profile all trajectories in the LIBERO training demonstrations across the four task suites. 
Fig.~\ref{fig:RnD-distribution} visualizes the empirical distributions of $D^{[w]}$ and $R^{[w]}$ for each suite, revealing notable differences in their ranges and tail behaviors.
As shown in the Fig.~\ref{fig:RnD-distribution}, both indicators exhibit long-tailed distributions with occasional extreme outliers caused by sudden motions or annotation artifacts. 
Directly using the global maximum as the upper bound would compress the majority of valid samples into a narrow range, reducing discriminative power.

\subsection{Min-Max Normalization with Percentile Clipping}

To address this issue, we adopt a \textbf{min-max normalization} scheme with \textbf{95$^{\textnormal{th}}$ percentile clipping}. Specifically, for each indicator $x \in \{D^{[w]}, R^{[w]}\}$ and task suite $s$, we compute:
\begin{equation}
x_{\min}^{(s)} = \min_{i} x_i^{(s)}, \quad x_{\max}^{(s)} = Percentile_{\textnormal{95}}\left(\{x_i^{(s)}\}\right),
\end{equation}
where $x_i^{(s)}$ denotes the $i$-th sample value in suite $s$. 
The normalized indicator is then computed as:
\begin{equation}
\hat{x} = clip\left(\frac{x - x_{\min}^{(s)}}{x_{\max}^{(s)} - x_{\min}^{(s)}}, \ 0, \ 1\right),
\label{eq:minmax-norm}
\end{equation}
where the $clip(\cdot, 0, 1)$ function ensures that values exceeding the 95$^{\textnormal{th}}$ percentile threshold are capped at 1, and values below the minimum are capped at 0.

This design choice is motivated by two observations: (1) the 95$^{\textnormal{th}}$ percentile effectively excludes extreme outliers while preserving the dynamic range of typical trajectories, and (2) capping the normalized output to $[0, 1]$ provides a well-bounded input for threshold-based switching decisions.

\begin{table}[!b]
\centering
\footnotesize
\caption{Normalization bounds for $D^{[w]}$ and $R^{[w]}$ across LIBERO task suites. All values are derived from the training demonstrations, with the upper bound set at the 95$^{\textnormal{th}}$ percentile.}
\label{tab:norm-bounds}
\begin{tabular}{c|c|c|c|c}
\toprule
\toprule
\textbf{Task Suite} & $D_{\min}$ & $D_{\max}$ (95\%) & $R_{\min}$ & $R_{\max}$ (95\%) \\
\midrule
LIBERO-Goal    & 0.000009 & 0.123381 & 0.000001 & 0.014989 \\
LIBERO-Spatial & 0.000027 & 0.128629 & 0.000019 & 0.015654 \\
LIBERO-Object  & 0.000098 & 0.116458 & 0.000010 & 0.014151 \\
LIBERO-Long    & 0.000008 & 0.102298 & 0.000001 & 0.012479 \\
\bottomrule
\bottomrule
\end{tabular}
\end{table}

\subsection{Suite-Specific Normalization Bounds}

Based on our profiling, the suite-specific normalization bounds are summarized in Tab.~\ref{tab:norm-bounds}.
These bounds are applied at inference time to normalize raw kinematic values before comparing them against the switching thresholds. 
By using suite-specific bounds, we ensure that the adaptive switching mechanism remains calibrated to the characteristic motion patterns of each task category.

% ==================================================================
% section: Real World Evaluation Details
% ==================================================================
\section{Real-World Evaluation Details}
\label{apdx:RealWorldExpr}

\subsection{Tabletop Operation Environment}
\label{apdx:RealWorldExpr-Tabletop}

\begin{figure}[!t]
  \begin{center}
    \centerline{\includegraphics[width=6.7in]{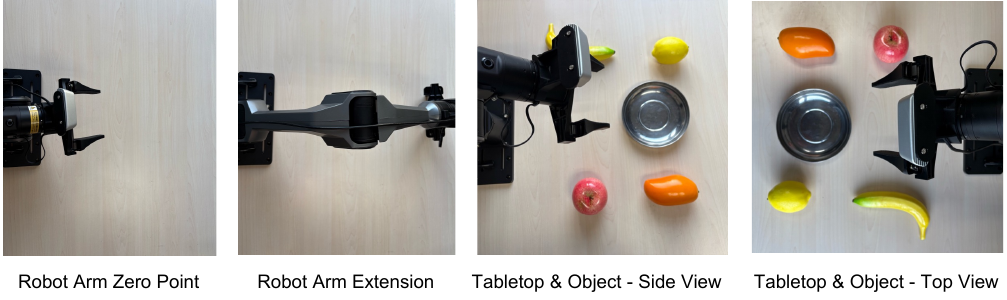}}
    \caption{Our Tabletop Operation Environment.}
    \label{fig:TabletopEnv}
  \end{center}
\end{figure}

% 桌面实验环境怎么建立的? (描述一下)
We build a tabletop operation environment for real-world experiments, as depicted in Fig.~\ref{fig:TabletopEnv}.
Specifically, we fix the robotic arm on a common tabletop (located at the midpoint of the tabletop), ensuring that the operating range of the robotic arm can cover the entire tabletop.
We use the official tabletop fixer to lock the position of the robotic arm.
Moreover, we use a variety of objects as manipulation target, including foam fruit models and kitchen utensils (e.g., a stainless steel plate).

\subsection{Robot Arm}
\label{apdx:RealWorldExpr-RobotArm}
\begin{figure}[!b]
  \begin{center}
    \centerline{\includegraphics[width=6.7in]{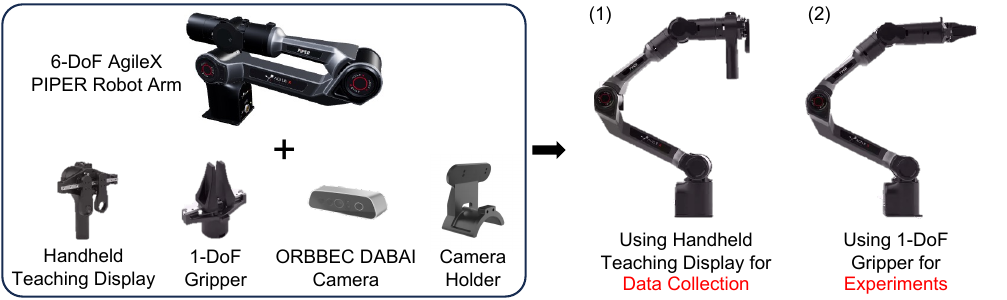}}
    \caption{Details of the Robot Arm.}
    \label{fig:RobotArm}
  \end{center}
\end{figure}

We use a popular 6-DoF AgileX PIPER robotic arm.
Its accessories include a handheld teaching display device, a 1-DoF gripper, an ORBBEC DABAI camera, and a plastic camera holder, as shown in Fig.~\ref{fig:RobotArm}.
There are two solutions for the assembly of the robotic arm: (1)
When the Handheld Teaching Display Device is mounted on the robotic arm, it can be used for data collection. (2) When the gripper is mounted on the robotic arm, it can be used for real-world experiments.

\subsection{Task Establish}
\label{apdx:RealWorldExpr-TaskEstab}
We set up several tasks based on the tabletop operating environment.
Considering that existing robotic arm datasets such as RoboCasa~\cite{robocasa} are confined to simulation, and while the Open X-Embodiment (OXE) dataset~\cite{OXE} offers a large scale, it suffers from hardware heterogeneity and is difficult to reproduce in a single laboratory setting, whereas the tasks in SimplerEnv~\cite{simplerenv} are overly simplistic. 
We integrate mainstream tasks from these datasets, such as object manipulation (grasping and moving) and placing objects into designated containers. 
Therefore, we introduce diverse fruit models and containers, along with environmental variations (e.g., lighting conditions and backgrounds), to facilitate the execution of diverse tasks within a real-world tabletop operation environment.
A comparison between our tasks and existing datasets is presented in Table~\ref{tab:dataset_comparison}.

\input{tab/apdx-taskcompare.tex}

Specifically, the established tasks encompass simple grasping tasks (e.g., \texttt{pick up the apple/banana}), spatial displacement tasks (e.g., \texttt{move the apple from point A to point B}), and specific pick-and-place tasks (e.g., \texttt{pick up the mango from the plate and place it into the bowl}). Furthermore, we include more complex composite tasks, such as \texttt{removing the apple from the plate and placing it into the bowl, then placing the banana onto the plate}. 
To account for the complexity of real-world environments, we conducted operations under varying lighting conditions, fruit categories, and container styles, as detailed in Table~\ref{tab:task_summary}.

\input{tab/apdx-TaskSummary.tex}

\subsection{Data Collection}
\label{apdx:RealWorldExpr-DataCollect}
We constructed the dataset using Google's Robot Learning Dataset Specification (RLDS), encapsulating it in a hierarchical and serialized HDF5 format. 
The data collection platform is based on a PIPER 6-DoF robotic arm (communicating via CAN bus) and equipped with an ORBBEC DABAI camera for primary-view visual feedback. 
Regarding data processing and state representation, the system synchronously collects visual and proprioceptive data at 10 Hz. 
Visual observations are resized from the original $640 \times 480$ resolution to $224 \times 224$ pixels via bilinear interpolation and converted to the RGB color space to adapt to model inputs. 
The robot state vector consists of six joint angles (unified to radians) and a binary gripper state, where the open/close status is automatically determined via an adaptive threshold algorithm based on the difference between initial calibration and real-time feedback. 
The action space is defined as the relative increments (delta joint positions) of joint angles between adjacent time steps and the absolute state of the target gripper. 
Regarding the collection protocol, each session begins with the robotic arm's automatic enabling and zero calibration. 
Subsequently, the operator inputs a natural language task instruction and switches to Teach Mode to complete object transport tasks by manually manipulating the arm joints and controlling the gripper via a teach pendant.
Each episode contains complete sequential image, state, action, and timestamp information, with task instructions and outcome-based sparse rewards (1.0 for success, -1.0 for failure) recorded in the file metadata. 
To enhance data robustness, repeated trials were conducted under various lighting and background conditions to increase diversity, collecting approximately 300 episodes per task type for subsequent fine-tuning.

\subsection{Model Fine-Tuning}
\label{apdx:RealWorldExpr-ModelFineTune}
% 基于什么做微调？PEFT
We employed a Parameter-Efficient Fine-Tuning (PEFT)~\cite{PEFT-1, PEFT-2, PEFT-3} strategy based on the OpenVLA-7B model clusters. 
Specifically, utilizing Low-Rank Adaptation (LoRA)~\cite{lora-1, lora-2}, we trained by injecting low-rank adapters with a rank of $r=32$ and a dropout rate of $0.05$ while freezing the pretrained backbone parameters; during inference, task-specific statistics are dynamically loaded to achieve action space un-normalization.
Regarding inference deployment, the system adopts a Client-Server (C/S) architecture, with the computing end acting as the server and the robotic arm as the client. 
The server loads the model using bfloat16 half-precision and the Flash Attention 2 acceleration mechanism. 
It encapsulates client-uploaded images and natural language instructions (e.g., "pick up the banana") into a Q\&A prompt template, employing a greedy decoding strategy to predict a 7-dimensional action vector. 
Notably, addressing the multi-threading characteristics of HTTP, we introduced a synchronization and mutual exclusion mechanism based on condition variables at the inference interface. 
Through the cooperative control of a global status flag and a condition lock, this mechanism establishes the core inference process as a critical section, enforcing the serialized execution of requests. 
This design effectively eliminates risks of GPU memory overflow and resource competition caused by concurrent multi-threaded GPU calls, while maintaining high-concurrency network communication capabilities. 
Upon receiving the action values (relative joint increments) from the server, the client converts them into absolute target angles via an integration algorithm and determines the gripper's open/close status. 
Finally, the results are mapped to SDK underlying pulse values to drive the PIPER robotic arm via the CAN bus, achieving high-precision closed-loop control.

%% file: tab/apdx-LIBEROstat.tex
\begin{table}[!b]
\centering
\footnotesize
\vskip -0.1 in
\caption{LIBERO Dataset Statistics}
\label{tab:dataset_stats}
\begin{tabular}{c|c|c|c|c|c|c}
\toprule
\toprule
\textbf{Dataset} & \textbf{Tasks} & \textbf{Episodes} & \textbf{Total Steps} & \textbf{Avg. Steps/Episode} & \textbf{Size} & \textbf{Files} \\
\midrule
LIBERO-Goal     & 10 & 428  & 52,042  & 121.59 $\pm$ 37.62 & 1.7 GB & 18 \\
LIBERO-Spatial  & 10 & 432  & 52,970  & 122.62 $\pm$ 20.75 & 1.8 GB & 18 \\
LIBERO-Object   & 10 & 454  & 66,984  & 147.54 $\pm$ 18.53 & 2.6 GB & 34 \\
LIBERO-Long     & 10 & 379  & 101,469 & 267.73 $\pm$ 56.74 & 3.4 GB & 34 \\
\bottomrule
\bottomrule
\end{tabular}
\end{table}

%% file: tab/apdx-DBdetails.tex
\begin{table}[!t]
\centering
\footnotesize
\caption{Vector Database Storage Statistics}
\label{tab:db_stats}
\begin{tabular}{c|c|c|c}
\toprule
\toprule
\textbf{Dataset} & \textbf{Ttl. Vectors} & \textbf{Size} & \textbf{Ttl. Size}\\
\midrule
LIBERO-Goal    & 52,042  & 1.38 GB  & \multirow{4}{*}{6.5 GB}\\
LIBERO-Object  & 66,984  & 1.58 GB & \\
LIBERO-Spatial & 52,970  & 1.38 GB & \\
LIBERO-Long    & 101,469 & 2.16 GB & \\
\bottomrule
\bottomrule
\end{tabular}
\vskip -0.2 in
\end{table}

%% file: tab/apdx-taskcompare.tex
\begin{table}[!t]
\centering
\footnotesize
\caption{Task Comparison with Existing Popular Datasets}
\begin{tabular}{c|c|c|c|c}
    \toprule
    \toprule
    % 表头
    \textbf{Dataset} & \textbf{Domain} & \textbf{Hardware Setup} & \textbf{Task Complexity} & \textbf{Obj. \& Env. Diversity} \\
    \midrule
    OXE~\cite{OXE} & Real & Heterogeneous & Atomic / Short-horizon & High \\
    RoboCasa~\cite{robocasa} & Simulation & Standardized & Long-horizon & High \\
    SimplerEnv~\cite{simplerenv} & Sim / Real & Standardized & Atomic & Low \\
    \textbf{Ours} & \textbf{Real} & \textbf{Standardized} & \textbf{Hierarchical} & \textbf{High} \\
    \bottomrule
    \bottomrule
  \end{tabular}
\label{tab:dataset_comparison}
\end{table}

%% file: tab/apdx-TaskSummary.tex
\begin{table}[!t]
  \centering
  \footnotesize
  \caption{Task Categories and Examples}
  \label{tab:task_summary}
  \renewcommand{\arraystretch}{1.5}
  \setlength{\tabcolsep}{6.7pt}
  \resizebox{\textwidth}{!}{
  \begin{tabular}{l | p{6cm} | p{5.5cm} | c}
    \toprule
    \toprule
    \textbf{Task Category} & \textbf{Description} & \textbf{Instruction Examples} & \textbf{Complexity} \\
    \midrule
    Atomic Grasping & 
    Grasping a specific target object from the tabletop without interacting with other objects. & 
    $\bullet$ ``Pick up the apple.'' \newline 
    $\bullet$ ``Pick up the banana.'' & 
    Low \\
    \cmidrule{1-4}
    Spatial Displacement & 
    Moving an object from a starting position to a target region or a specific container. & 
    $\bullet$ ``Move the apple from point A to point B.'' \newline 
    $\bullet$ ``Pick up the mango from the plate and put it into the bowl.'' & 
    Medium \\
    \cmidrule{1-4}
    Composite Sequential & 
    Long-horizon tasks requiring multi-step planning and state memory to manipulate multiple objects in sequence. & 
    $\bullet$ ``Take the apple out of the plate and put it in the bowl, then place the banana onto the plate.'' & 
    High \\
    \bottomrule
    \bottomrule
  \end{tabular}
  }
\end{table}